\documentclass[letterpaper, 10 pt, journal, twoside]{ieeetran}  

\usepackage{cite}
\usepackage{xcolor}
\usepackage{graphicx}
\usepackage{amsmath}

\usepackage{enumitem}
\usepackage{array}
\usepackage{tabularx}
\usepackage{makecell}
\usepackage{dirtree}
\usepackage{multirow}
\usepackage{booktabs}
\usepackage[utf8]{inputenc}
\usepackage{subcaption}
\usepackage{url}

\IEEEoverridecommandlockouts                              

\title{CAVERS: Multimodal SLAM Data from a Natural Karstic Cave with Ground Truth Motion Capture}

\author{
Giacomo Franchini$^{1}$, 
David Rodríguez-Martínez$^{2}$,
Alfonso Martínez-Petersen$^{2}$, \\
Carlos~J.~Pérez-del-Pulgar$^{2}$ and Marcello Chiaberge$^{1}$
\thanks{Manuscript received: April, 16, 2026; Revised July, 11, 2026; Accepted August, 7, 2026.}
\thanks{This paper was recommended for publication by Editor Ayoung Kim upon evaluation of the Associate Editor and Reviewers’ comments.} 
\thanks{$^{1}$Giacomo Franchini and Marcello Chiaberge are with Polytechnic of
         Turin Interdepartmental Centre for Service Robotics (PIC4SeR), Corso
         Francesco Ferrucci 112, 10141 Turin, Italy (emails: \tt\footnotesize
         giacomo.franchini@polito.it, marcello.chiaberge@polito.it)}
 \thanks{$^{2}$David Rodríguez-Martínez, Alfonso Martínez-Petersen, and Carlos Jesús Pérez-del-Pulgar are with Systems Engineering and Automation Department, Universidad de Málaga, Ortiz Ramos s/n, 29010, Málaga, Spain (emails: \tt\footnotesize david.rm@uma.es, alfonsomp@uma.es, carlosperez@uma.es)}
 \thanks{Digital Object Identifier (DOI): see top of this page.}
}

\begin{document}

\maketitle

\begin{abstract}
Autonomous robots operating in natural karstic caves face perception and navigation challenges that are qualitatively distinct from those encountered in mines or tunnels: irregular geometry, reflective wet surfaces, near-zero ambient light, and complex branching passages. Yet publicly available datasets targeting this environment remain scarce and  
offer limited sensing modalities and environmental diversity.
We present CAVERS, a multimodal dataset acquired in two structurally distinct rooms of Cueva de la
Victoria, Málaga, Spain, containing 24 sequences with approximately 335 GB of recorded data. 
The sensor suite combines an Intel RealSense D435i RGB-D-I camera, an Optris PI640i near-IR thermal camera, and a Velodyne VLP-16 LiDAR, operated both handheld and mounted on a wheeled rover under full darkness and artificial illumination. For most of the sequences, mm-accurate 6-DoF ground truth pose and velocity at 120 Hz are provided by an Optitrack motion capture system installed directly inside the cave. We benchmark seven state-of-the-art SLAM and odometry algorithms spanning visual, visual-inertial, thermal-inertial, and LiDAR-based pipelines, as well as a 3D reconstruction pipeline, demonstrating the dataset's usability. The dataset and all supplementary material are publicly available at: \url{https://github.com/spaceuma/cavers}.
\end{abstract}


\section{Introduction}
\label{sec1_introduction}
\IEEEPARstart{U}{nderground} cavities, such as caves and karst formations, are of enduring scientific value. They host unique biological organisms \cite{intro::caves::culver,intro::caves::kosznik}, preserve geological and climatic records otherwise lost at the surface, and supply subsurface water and mineral resources \cite{intro::caves::ford}. However, many of these subterranean environments are high risk or inaccessible to humans both on Earth and beyond. On the Moon and Mars, multiple skylights have been identified as potential entries to a network of underground lava tubes \cite{kaku2017detection}, which are now priority targets for exploration. Whether for driving scientific characterization, supporting search and rescue operations, or scouting planetary caves, robots offer a way to explore these environments where human access is either slow, dangerous, or unfeasible \cite{intro::cave_exp::tabib, intro::cave_exp::petracek}.

Autonomous robotic operations in underground environments are, however, particularly challenging. Darkness impairs passive vision, while feature-poor self-similar geometries degrade both visual and LiDAR odometry. Additionally, robots need to cope with incomplete or non-existent prior maps, unreliable communications, and the absence of any external positioning reference, such as GNSS. As a result, robots must rely solely on their on-board sensors for navigation in environments where dust, water, and thermal uniformity further degrade sensor reliability \cite{intro::dang}. Research in underground robotics has accelerated  following the DARPA Subterranean Challenge
\cite{intro::darpa}, which demonstrated that heterogeneous robotic systems can
autonomously explore large subterranean environments \cite{intro::darpa::cerberus}. While the challenge stimulated significant advances in Simultaneous Localization And Mapping (SLAM) in these environments, results also highlighted that reliable operation strongly depends on environmental structure, sensing configuration, and operational assumptions \cite{intro::darpa_slam}. Public datasets, therefore, play a
critical role in benchmarking SLAM methods under realistic conditions.

\begin{figure*}[tp]
  \centering
\smallskip
\smallskip
  \includegraphics[width=0.82\linewidth]{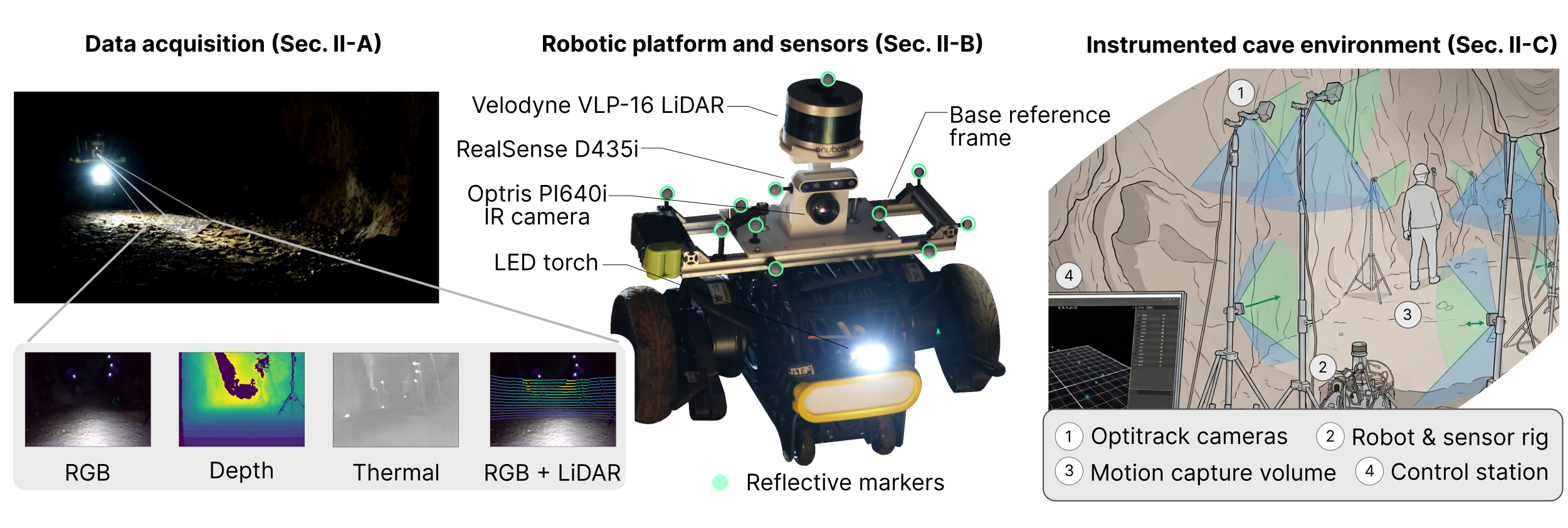}
  \caption{Dataset features overview: multimodal visual, range, and
  inertial data (Section~\ref{sec::2_2_sensors_calibration}) are collected using sensors operating in two configurations (Section~\ref{sec::2_1_seq_overview}). Ground truth pose and velocity of the
  sensor rig are provided by a motion capture system installed in the cave
  (Section~\ref{sec::2_3_ground_truth}).}
  \label{fig::sec1::dataset_overview}
  \vspace*{-10pt}
\end{figure*}

Most publicly available datasets focus on artificial underground environments such as
mines and tunnels \cite{intro::dataset::vli, intro::dataset::locus,
intro::dataset::chile,intro::dataset::min3d,intro::dataset::tunnel-collection}. These typically provide a combination of LiDAR, RGB, thermal, and inertial data, with varying robotic system locomotor topologies, from tracked and wheeled \cite{intro::dataset::subt-tunnel} to legged robots \cite{intro::darpa::cerberus}, or a combination thereof \cite{intro::dataset::subt-mrs}. Some works emphasized the perception challenges robots face when underground, such as high optical depth \cite{intro::dataset::multimodal-vdc} or high dynamic range when active illumination is used \cite{intro::dataset::oivio}. These datasets have enabled progress on SLAM in structured subterranean environments. However, mines and tunnels present relatively regular geometries and predictable structures, which do not fully capture the complexity of natural caves. A smaller body of work targets natural environments. Some datasets have been collected in lava tubes, also due to their interest in planetary exploration \cite{planetary_cave_exp}. LiDAR point clouds, inertial data, and platform kinematic/visual-inertial odometry measurements collected by a legged robot across the Lava Beds National Monument, in California, are included in \cite{intro::dataset::locus}. Other works \cite{intro::dataset::hidding, intro::dataset::nasa} focus on mapping natural lava tubes at Mount Etna National Park, Italy, and at Craters of
the Moon National Park, Idaho, respectively. While these works move beyond artificial tunnels, lava tubes are characterized by a relatively regular, basaltic tunnel-like morphologies that differ substantially from karstic caves, which feature irregular geometries, narrow passages, vertical structures, high branching, and reflective surfaces due to water presence. These characteristics pose additional challenges for localization, mapping, and loop-closure detection. Only a few datasets explicitly target SLAM in natural caves: \cite{intro::dataset::subt-mrs} includes cave sequences recorded at the Laurel Caverns State Park, Pennsylvania, consisting of LiDAR scans, fisheye RGB images, and IMU data from a handheld platform equipped with an active light source, while in \cite{intro::cave_exp::petracek} the authors published the LiDAR-inertial data collected during a UAV flight inside the Bull Rock Cave Systems in Czech Republic. Both mainly focus on LiDAR-based navigation and remain limited in terms of sensing modalities and platform configurations. Moreover, these datasets estimate the ground truth trajectories by registering the on-board LiDAR scans with multiple cave point clouds obtained from a 3D scanner at different locations. Despite the ease of setup of the scanner, this indirect method requires additional steps, such as a first manual registration of the scans or the implementation of a loop closure pipeline, followed by a successive finer ICP-based alignment, which could lead to larger estimation errors for the ground truth trajectories.

To address these gaps, we introduce CAVERS, a multimodal dataset collected at Cueva de la Victoria in Málaga, Spain. The dataset includes synchronized RGB-D images, inertial data, rotating 3D LiDAR scans, and near-IR thermal sensing, acquired under different illumination conditions (full darkness or artificially illuminated) and platform configurations (handheld and wheeled rover). Unlike most published works, our dataset includes high-frequency, high-accuracy ground truth trajectories using an Optitrack motion-capture system installed in the cave, enabling a precise evaluation of SLAM performance. The data are provided in both rosbag format and as raw sensor outputs. An overview of the dataset features is reported in Figure~\ref{fig::sec1::dataset_overview}. The main contributions of our work are the following:
\begin{itemize}
  \item A multimodal dataset targeting SLAM in natural karstic caves, including almost 55,000 RGB, depth and thermal images and totaling 335 GB of recorded data;
  \item High-frequency mm-accurate ground truth pose and velocity for most sequences; 
  \item A comprehensive evaluation of state-of-the-art visual and LiDAR-based SLAM
  algorithms and testing of a 3D reconstruction pipeline on the collected data.
\end{itemize}

\section{The Dataset}
\label{sec2_dataset}
This article presents the CAVERS dataset, collected in two distinct rooms of the Cueva de la Victoria, one of the numerous natural cavities
in the province of Málaga, Spain. This cave is part of a larger complex named
Cuevas del Cantal, situated in the El Cantal cliffs, a limestone rocky
promontory between the city of Málaga and the nearby locality of Rincón de la
Victoria. Formed during the Jurassic age, El Cantal
has been affected by karst dissolution driven by the infiltration of both
continental and marine waters, which, over time, generated the caves' complex.
Two wells provide access to the cave: the first leads to a room called Sala del
Dosel, about 30 m long, with a maximum width of 5 m, and an average height of 5 m.
Cavities open on the room's right wall, leading to more than 150 m of tunnels,
up to a second room known as Sala de las Conchas. This room, measuring 50 m in
length, and an average of 6x2 m in width and height, is the entrance point of
the second well. The differences in geometry, scale, and structural complexity
of the two rooms provide diverse conditions for SLAM evaluation. Bringing all
the necessary equipment into the cave posed significant logistical challenges,
as sensitive hardware had to be transported through restricted passages and
uneven terrain. In particular, installing and calibrating the motion capture
system required careful planning to ensure optimal sensor placement and reliable
operation in a space-constrained environment and difficult lighting conditions.

\subsection{Sequences overview}
\label{sec::2_1_seq_overview}

\begin{table*}[t]
\smallskip
\smallskip
\renewcommand{\arraystretch}{0.9}
\caption{
  Overview of the trajectories, grouped by testing objective (\textit{loc} for localization and \textit{rec} for reconstruction) and sensor setup. 
  Indices in brackets identify the specific sequence within each group. If no indices are reported, the cell components apply to the entire trajectory group.
}
\label{tab::sec2::sequences}
\centering
\begin{minipage}{\textwidth}
  \begin{tabular}{c|c|c|c|>{\raggedright\arraybackslash}m{5.5cm}}
  \hline
  \textbf{Name\_\{traj\}}                       & \textbf{Room}         & \textbf{\{traj\}: Illumination}                                               & \textbf{\{traj\}: Ground truth}                                     & \multicolumn{1}{c}{\textbf{Notes}}\\
  \hline\hline

  \textit{loc\_diablo\_\{1-8\}}                 & Sala del Dosel        & \makecell{{\tiny\{1,2,3,5,6,8\}}: LED torch\\ {\tiny\{4,7\}}: No lights}         & 6D Pose and Velocity                                                &
    Complete traverses of the room. High motion blur and vibrations. Rock obstacles on the ground. \\
  \hline
  \textit{loc\_handheld\_\{1-4\}}               & Sala del Dosel        & \makecell{{\tiny\{1,4\}}: LED torch\\ {\tiny\{2,3\}}: No lights}                    & 6D Pose and Velocity                                                &
    Complete traverses of the room. Low speed, smooth omnidirectional motion. Variable sensors pointing.\\
  \hline
  \textit{loc\_handheld\_\{5-6\}}               & Sala de las Conchas   & \makecell{{\tiny\{5\}}: LED torch\\ {\tiny\{6\}}: No lights}                          & None                                                                &
    Complete traverse of the room. Low speed, smooth omnidirectional motion. Variable sensors pointing.\\
  \hline
  \textit{rec\_diablo\_\{1-2\}}                 & Sala del Dosel        & Cave lighting                                                                 & None                                                                &
    Scan of the room terrain and lower walls. High motion blur and vibrations. Sensors pointing at the terrain.\\
  \hline
  \textit{rec\_handheld\_\{1-5\}}               & Sala del Dosel        & Cave lighting                                                                 & \makecell{{\tiny\{1\}}: 6D Pose and Velocity\\ {\tiny\{2,3,4,5\}}: None}     &
    Scan of the room, including lateral walls and ceiling, with multiple passes on the same areas. Low speed, smooth motion.\\
  \hline
  \textit{rec\_handheld\_\{6-8\}}               & Sala de las Conchas   & \makecell{{\tiny\{6,7\}}: LED torch\\ {\tiny\{8\}}: No lights}                      & None                                                                &
    Scan of a cavity and of a vertical structure inside the rooms. Low speed, smooth motion.\\
  \hline
  \hline
  \end{tabular}
\end{minipage}
\vspace{-10pt}
\end{table*}

The dataset sequences are summarized in Table~\ref{tab::sec2::sequences}.
Data were collected by both operating the sensor
suite described in Subsection \ref{sec::2_2_sensors_calibration} in handheld
configuration and by installing it on a teleoperated Direct Drive Tech DIABLO
mobile wheeled platform. Handholding the sensor rig provides high motion
flexibility and allows complete scans of the cave’s rooms, including the roof
and walls, and the recording of data inside cavities that are more difficult for the robot to access. The DIABLO rover provides high payload stability and variable pitch
angles, enabling variation in sensor pointing during traverses and achieving
sensor motion and FOV resembling those observed during real robotic exploration
missions.

\begin{figure*}[tp]
\smallskip
\smallskip
  \centering
  \setlength\tabcolsep{2pt} 
  
  \begin{tabularx}{0.85\linewidth}{r c c c c}
    & RGB & Depth & Thermal & RGB + LiDAR \\
    
    \rotatebox{90}{\quad \textit{loc\_diablo\_1}} &
    \includegraphics[width=0.2\linewidth]{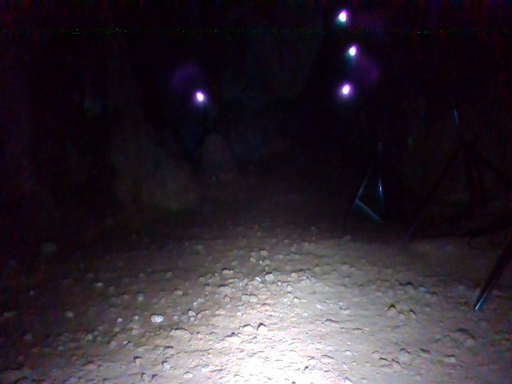} & 
    \includegraphics[width=0.2\linewidth]{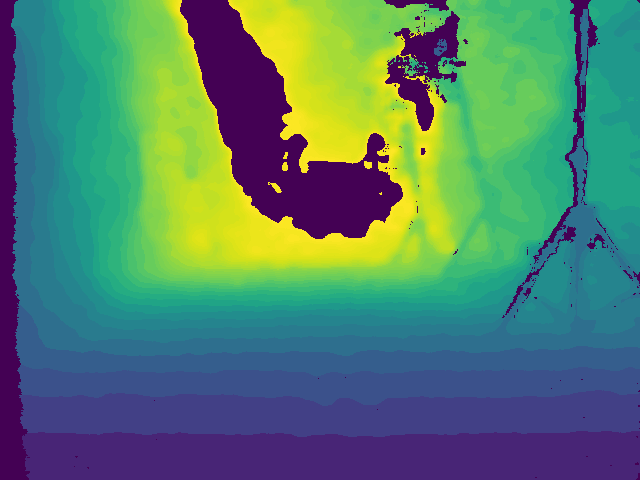} & 
    \includegraphics[width=0.2\linewidth]{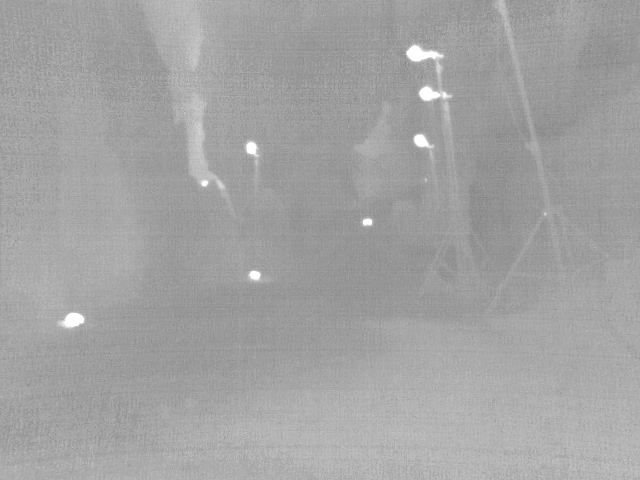} & 
    \includegraphics[width=0.2\linewidth]{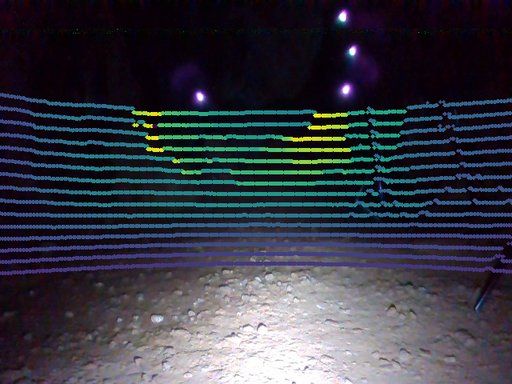} \\

    \rotatebox{90}{\quad \textit{loc\_handheld\_5}} &
    \includegraphics[width=0.2\linewidth]{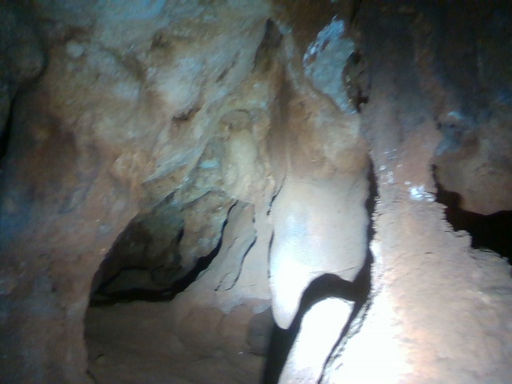} & 
    \includegraphics[width=0.2\linewidth]{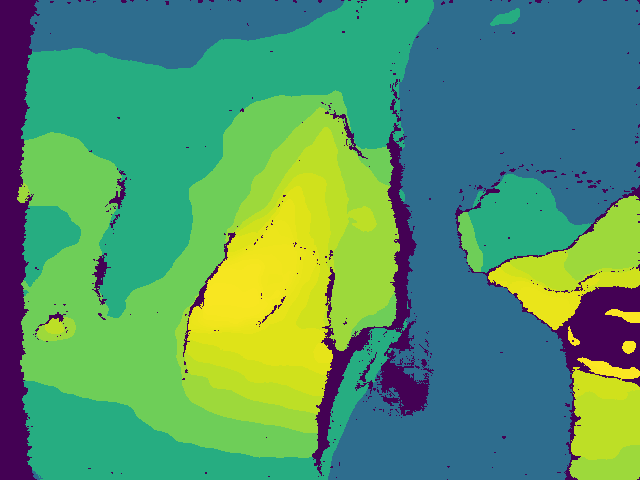} & 
    \includegraphics[width=0.2\linewidth]{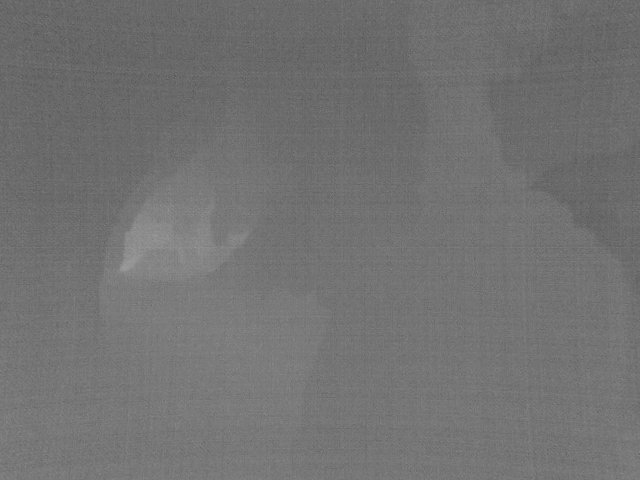} & 
    \includegraphics[width=0.2\linewidth]{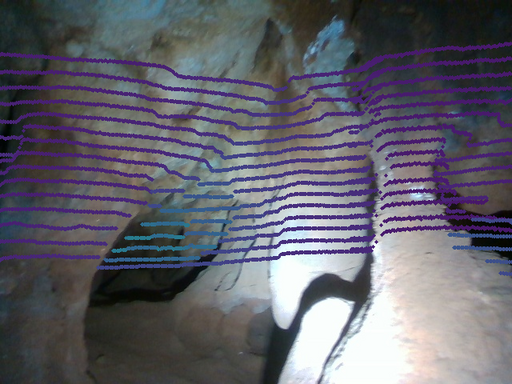} \\

    \rotatebox{90}{\quad \textit{rec\_handheld\_6}} &
    \includegraphics[width=0.2\linewidth]{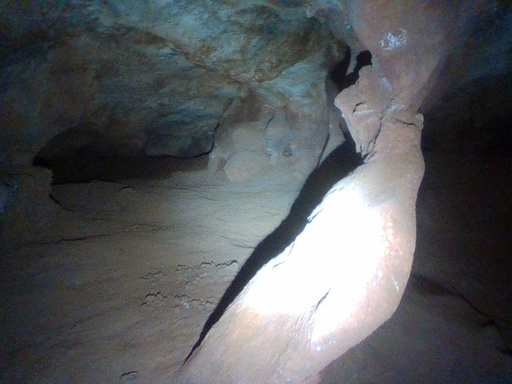} &
    \includegraphics[width=0.2\linewidth]{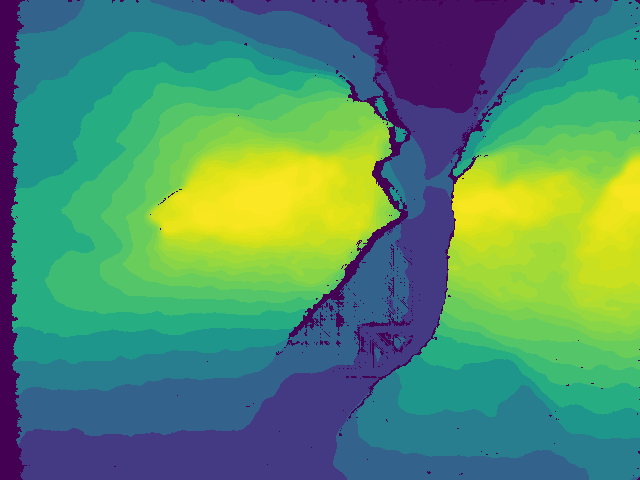} & 
    \includegraphics[width=0.2\linewidth]{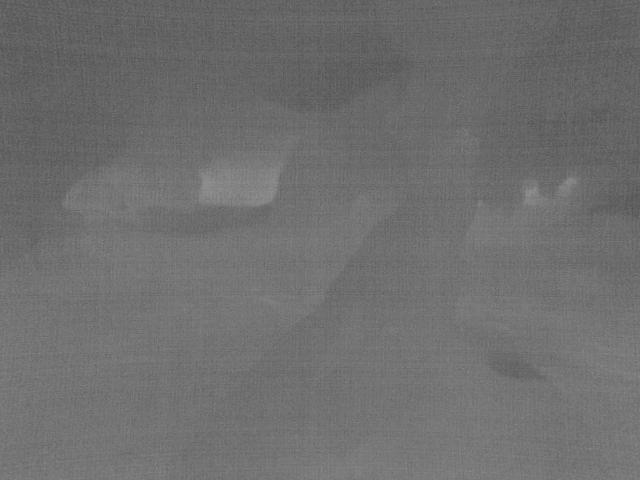} & 
    \includegraphics[width=0.2\linewidth]{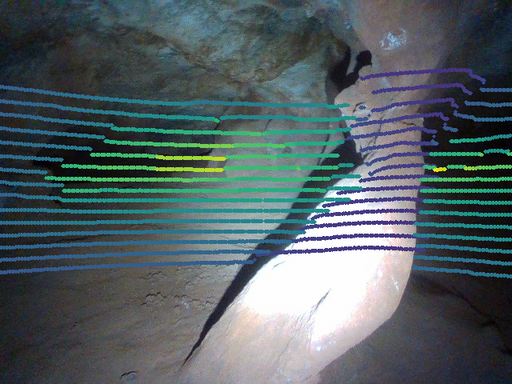} \\

    \rotatebox{90}{\quad \textit{rec\_handheld\_1}} &
    \includegraphics[width=0.2\linewidth]{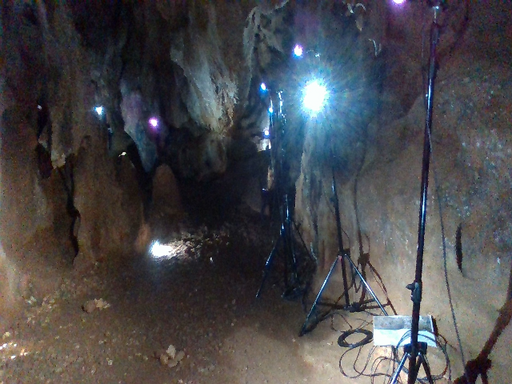} & 
    \includegraphics[width=0.2\linewidth]{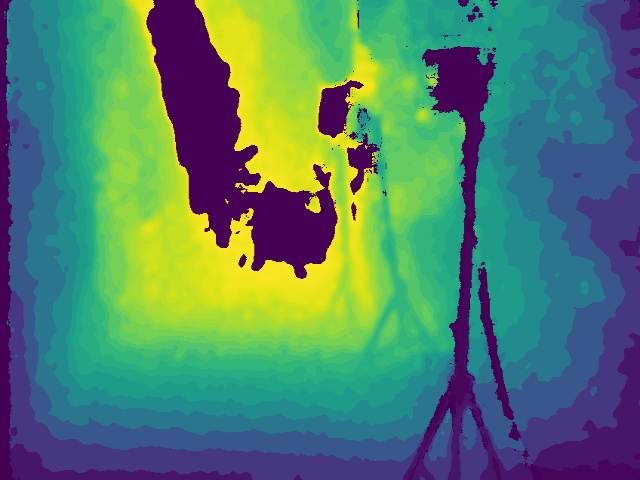} & 
    \includegraphics[width=0.2\linewidth]{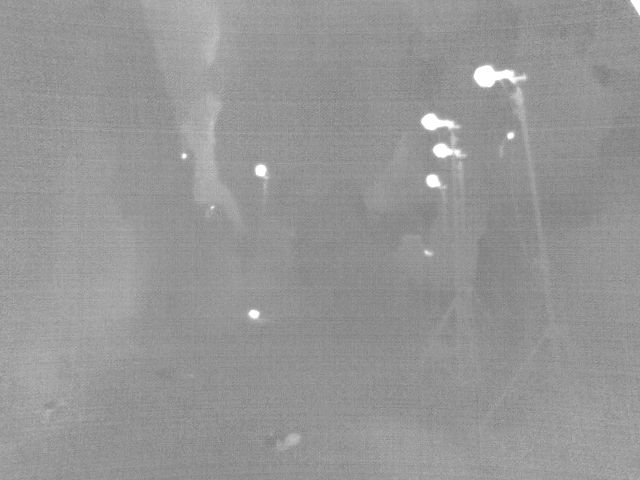} & 
    \includegraphics[width=0.2\linewidth]{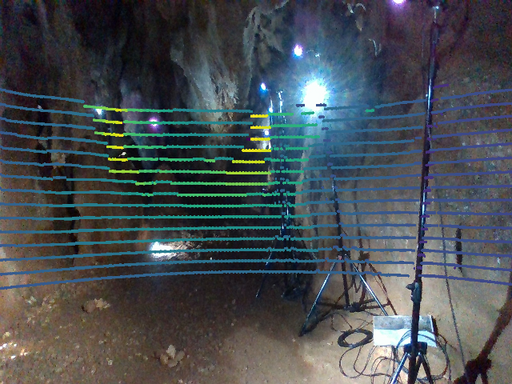} \\

    \rotatebox{90}{\quad \textit{loc\_handheld\_6}} &
    \includegraphics[width=0.2\linewidth]{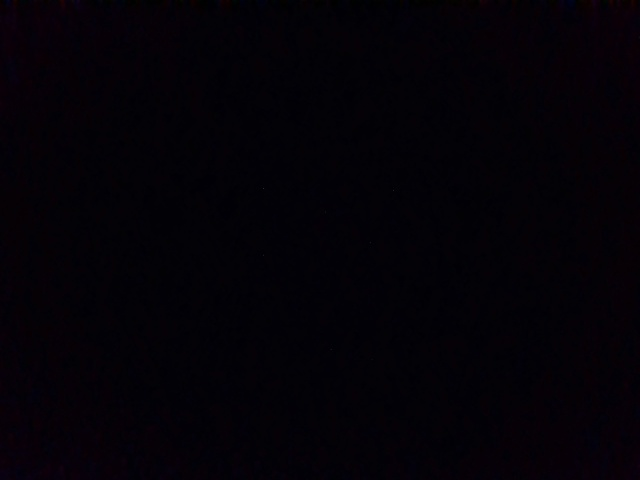} & 
    \includegraphics[width=0.2\linewidth]{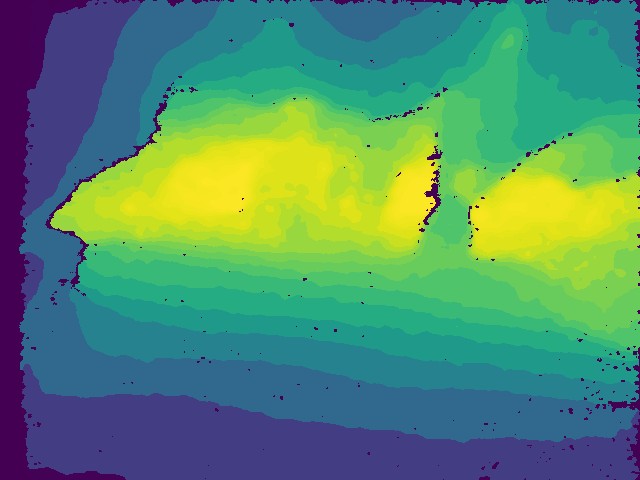} & 
    \includegraphics[width=0.2\linewidth]{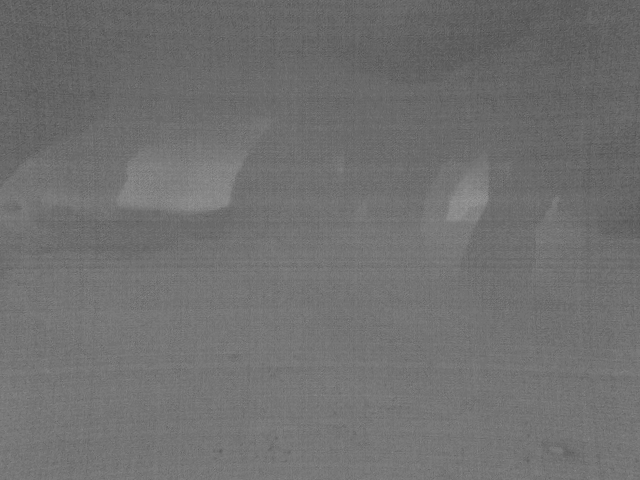} & 
    \includegraphics[width=0.2\linewidth]{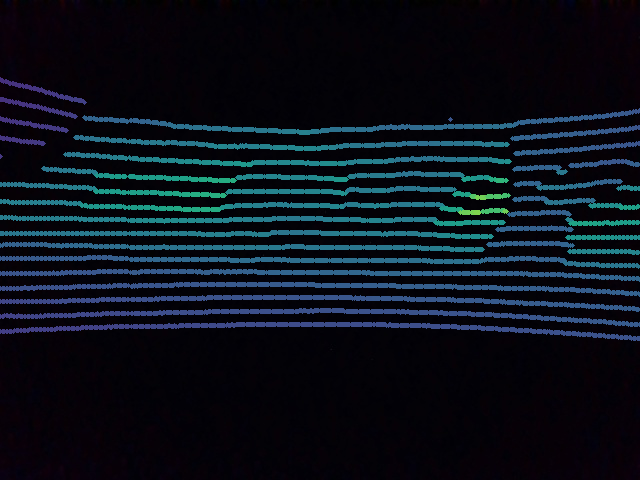} \\
  \end{tabularx}
  
  \caption{Comparison of sensor modalities across different lighting scenarios:
  LED torch (\textit{loc\_diablo\_1}, \textit{loc\_handheld\_5},
  \textit{rec\_handheld\_6}); cave lighting (\textit{rec\_handheld\_1}); no
  lights (\textit{loc\_handheld\_6}).}
  \label{fig::sec2::example_data}
\vspace*{-16pt}
\end{figure*}

The trajectories are recorded in variable illumination conditions, namely: sequences illuminated by an LED headlight mounted on the robot, and sequences recorded in total darkness. This stressed the sensor’s capabilities under different conditions. For example, we
occasionally observed missing depth values in areas directly illuminated by the
LED headlight, as shown in the third row of Figure~\ref{fig::sec2::example_data},
most likely because the RealSense camera could not perform stereo matching due
to the high reflectivity of the cave surfaces. The depth image is otherwise
complete when the same sequence is recorded in full darkness, but no
information from the RGB images is available in this case, leaving a
visual-based SLAM front-end module to rely solely on thermal and depth data. A
small subset of the sequences was recorded with the built-in cave artificial lighting system enabled to
facilitate mapping.

All sequences are designed to support SLAM research in natural
underground environments, but differ in their primary experimental
objectives and in the availability of ground truth data. Trajectories identified with the \textit{loc} prefix are intended for quantitatively evaluating the trajectory accuracy of SLAM and odometry
algorithms. For these, we provide ground truth values for the 6D pose and
velocity of the sensor rig obtained from an Optitrack motion capture system as
described in Section~\ref{sec::2_3_ground_truth}. Due to the impossibility of installing
the motion capture system inside Sala de las Conchas, ground truth is not
available for \textit{loc\_handheld\_5} and \textit{loc\_handheld\_6}
trajectories. However, visual methods can still be benchmarked by adopting the trajectories obtained from LiDAR-based SLAM algorithms as ground truth, using the provided scans, as shown in Section~\ref{sec::3_1_evaluation_slam}. Sequences identified with the \textit{rec} prefix are intended for 3D reconstruction and mapping of the cave rooms. Except for
\textit{rec\_handheld\_1}, here ground truth is not provided since the motion capture infrastructure was removed to enable an unobstructed view of the environment.

\begin{figure}[ht]
    \centering
    \includegraphics[width=0.85\linewidth]{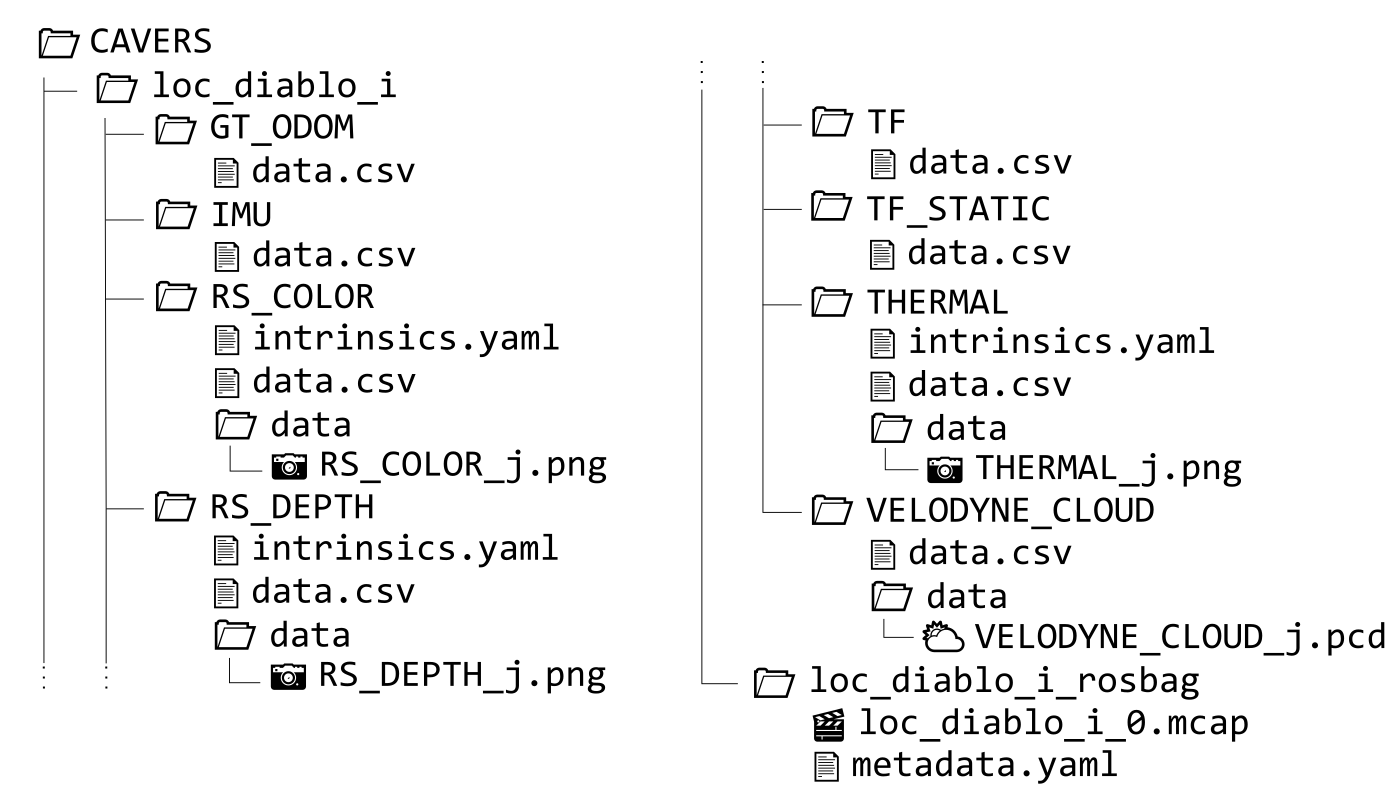}
    \caption{The dataset structure.}
    \label{fig::sec2::dataset_structure}
    \vspace*{-20pt}
\end{figure}

Data are provided as Robot Operating System 2 (ROS 2) rosbags in MCAP
format and also as a collection of extracted sensors and ground truth raw data, for
use in a ROS-agnostic way. RGB images are provided in 3-channel, 8-bit, lossless
PNG format. Depth images and thermal images are provided in single-channel,
16-bit, lossless PNG format. LiDAR point clouds are exported in PCD format.
Image and cloud names follow the convention \texttt{SENSORTYPE\_IDX.*}, with an
additional \textit{data.csv} file that relates the data index \texttt{IDX} with
the timestamp at which it has been recorded. IMU, ground truth, and
transformation data are all reported in CSV format with the associated
timestamps. A slice of the dataset structure is reported in
Figure~\ref{fig::sec2::dataset_structure}. 

\vspace{-5pt}

\subsection{Sensors, calibration and synchronization}
\label{sec::2_2_sensors_calibration}

\textbf{RGB-D-I camera} An Intel RealSense D435i camera provides RGB and depth
images. Images are delivered at VGA resolution (640x480p for both RGB and depth)
in SLAM evaluation sequences, to match the IR thermal camera's resolution and
simplify sensor fusion, and at HD resolution (1920x1080p for RGB, 848x480p for
depth) in cave reconstruction sequences, for maximum quality. The RGB camera
field of view (FOV) is 69°x42° in both resolutions, while the depth camera FOV
is 75°x62° and 87°x58° in VGA and HD resolution, respectively. Depth and RGB
images are streamed at 15 Hz. The camera is also equipped with an IMU providing
angular rates at 200 Hz and linear accelerations at 62 Hz. We deliver joined
gyroscope and accelerometer data at the gyroscope's frequency by enabling linear
interpolation of accelerations as provided by the official camera driver library
\textit{librealsense}.\\
\textbf{Thermal camera} The thermal footprint of the cave has been recorded with
an Optris PI640i near-IR thermal camera, equipped with a 60°x45° FOV lens
manually registered to focus objects at 1 m from the camera. The Optris camera
detects light in the 8--14 µm spectral range, with a temperature range from -20°C
to 100°C and a thermal sensitivity of 75 mK, expressed as the Noise Equivalent
Temperature Difference (NETD).
The data acquisition pipeline and formatting convention follow those of the
BASEPROD dataset~\cite{baseprod}, which employed the same sensor.
The camera outputs temperature values as a
VGA-resolution matrix of floats at approximately 6 Hz, which we provide in the dataset
as normalized, single-channel thermal images of 16-bit unsigned integers.
Normalization is performed by linearly scaling the float values $T_{\text{float}}$
within the fixed temperature interval $[T_{\text{min}}=14 \text{°} C; T_{\text{max}} =
20\text{°}C]$ as described in Equation~\ref{eq::sec2::thermal_image_rescale}. Raw float temperature matrices are available in the rosbags provided as
supplementary material.\\

\begin{equation}
\label{eq::sec2::thermal_image_rescale}
  T_{\text{uint16}} = \frac{T_{\text{float}} - T_{\text{min}}}{T_{\text{max}} - T_{\text{min}}} \cdot 65535
\end{equation}

\textbf{LiDAR} Lastly, our sensor setup includes a Velodyne VLP-16
360°-rotating LiDAR providing point cloud data with a vertical FOV of 30°. The
sensor has been configured in the strongest return mode with a maximum of 289344 points
per second, and with the angular speed set to its lowest value of 300 rpm to
achieve the finest azimuthal resolution of 0.1°. Point clouds extracted from the
sensors have been clipped to a distance range $\geq$ 0.1 m. We note the
presence of a dynamic obstacle behind the sensor, caused by the operator
following the rig. Raw sensor packets are available in the rosbags provided as
supplementary material.\\
\textbf{Calibration and synchronization} To determine the intrinsic and lens-distortion parameters
of the RGB and the IR thermal cameras, a series of images of a 12x9 checkerboard
target pattern with a glass background layer has been recorded. A collection of
calibration images for the thermal camera has been recorded by increasing the target
surface temperature with a heat source and exploiting the different irradiance
levels of black and white squares. The glass layer helps maintain a stable
temperature during image recording. Parameters are estimated with the MATLAB built-in
\textit{estimateCameraParameters} function, which is based on
\cite{calib::zhang2000} and \cite{calib::heikkila1997}. The RealSense IMU is calibrated with the official pipeline provided by Intel. This estimates the accelerometer biases
$\mathbf{b}_{acc}$ plus the sensitivity and off-axis terms described by the
matrix $\mathbf{S}$. For the gyroscopes, only the biases $\mathbf{b}_{gyro}$ are
obtained. The intrinsic parameters are written into the camera memory and used to
automatically calibrate the raw data using the models described in
Equation~\ref{eq::sec2::imu_calib}.

\begin{equation}
\label{eq::sec2::imu_calib}
\begin{aligned}
  \mathbf{a}_{calib} &= \mathbf{S} \mathbf{a}_{raw} - \mathbf{b}_{acc} \\
  \boldsymbol{\omega}_{calib} &= \boldsymbol{\omega}_{raw} - \mathbf{b}_{gyro}
\end{aligned}
\end{equation}

The transformation matrices from the rig's base frame to the IMU, LiDAR and
thermal camera frames are obtained from the assembly CAD model. The base reference
frame is defined to follow the ROS Enhancement Proposals (REP) 103 axis
orientation convention, with $x$ axis pointing forward, $y$ axis pointing left,
and $z$ axis pointing up. Its origin is conventionally placed in the rear-left
corner of the rig's support plate as displayed in Figure~\ref{fig::sec1::dataset_overview}. The RealSense D435i provides factory-calibrated extrinsic parameters between the internal IMU, RGB camera, and the
left IR projector frames, which is the frame used as a reference for the depth images.




Sensors and ground truth data streams were logged using a unified temporal reference, effectively eliminating clock drift. However, limited internal delays may exist due to data processing and recording overhead. No hardware-level synchronization has been implemented, so timestamps are independent for each sensor.

\subsection{Ground truth}
\label{sec::2_3_ground_truth}

We provide ground truth values for the 6D pose and velocity of the sensor rig
for a subset of the dataset sequences. Ground truth poses are obtained from an
Optitrack motion capture system, composed of ten IR cameras installed in the
cave, providing tracking data at approximately 120 Hz. The motion capture system has been calibrated inside the cave by collecting approximately 10,000 points for each camera with a calibration wand. Calibration results are provided as Mean Ray Error and Mean Wand Error, representing the accuracy of the 3D point estimation and the difference between the detected and the expected calibration wand length, respectively. After calibration, both values are approximately 5 mm.
Ground truth pose and velocity
at time \textit{i} are provided as the tuples $(\mathbf{p}_i, \mathbf{q}_i)$ and
$(\mathbf{v}_i, \boldsymbol{w}_i)$, where $\mathbf{p}_i$ and $\mathbf{q}_i$
represent the rig's 3D position and orientation, the latter parametrized with
unit quaternion in scalar-last format, while $\mathbf{v}_i$ and $\boldsymbol{w}_i$ represent the
3D linear and angular velocities of the rig, respectively. Poses are provided in
the motion capture base reference system, which is fixed and set after
calibration. The base frame, named \textit{map}, has the origin placed at ground
level and the \textit{z} axis pointing up. Velocities are computed by
discrete-time differentiation of consecutive ground-truth poses and provided in
the sensor rig base frame.

\section{Evaluation}
\label{sec3_evaluation}
\begin{figure*}[tp]
\smallskip
\smallskip
    \centering
    
    \captionof{table}{Trajectory translation and rotation ATEs (m and °, respectively). Lowest errors are highlighted separately for visual and LiDAR methods. The suffixes D and I denote depth and inertial modalities. The symbol × indicates tracking loss without recovery, while – denotes that no error is computed because RTAB-Map LiDAR-I is the reference solution.}
    \label{tab::sec3::results}
    
    \renewcommand{\arraystretch}{0.9}
    \setlength{\tabcolsep}{2pt}

    \begin{tabular*}{\textwidth}{@{\extracolsep{\fill}}l cc cc cc}
    \hline
    \multirow{2}{*}{\centering\textbf{Algorithm}} & \multicolumn{2}{c}{\textbf{loc\_handheld\_1}}       & \multicolumn{2}{c}{\textbf{loc\_diablo\_5}}       & \multicolumn{2}{c}{\textbf{loc\_handheld\_5}} \\
    \cmidrule(lr){2-3} \cmidrule(lr){4-5} \cmidrule(lr){6-7}
                                                  & $ATE_{t}$       & $ATE_{rot}$                       & $ATE_{t}$       & $ATE_{rot}$                     & $ATE_{t}$         & $ATE_{rot}$           \\
    \hline
    \hline
    ORBSLAM3 RGB-D                                & x               & x                                 & x               & x                               & 0.897            & 5.329                 \\
    ORBSLAM3 RGB-D-I                              & 1.446           & 9.689                             & 1.214           & 18.114                          & \textbf{0.743}   & \textbf{2.963}        \\
    RTAB-Map RGB-D-I                              & 0.453           & 5.811                             & \textbf{0.628}  & 9.962                           & 0.828            & 10.595                \\
    ROVTIO                                        & \textbf{0.245}  & \textbf{3.485}                    & 5.530           & \textbf{8.91}                   & 2.105            & 6.250                 \\
    DROID-SLAM RGB-D                              & 0.485           & 17.379                            & 0.917           & 44.486                          & 1.397            & 31.322                \\
    \hline
    KISS-ICP                                      & 0.168           & 4.701                             & 0.224           & 6.466                           & \textbf{0.101}   & 1.578                 \\
    GENZ-ICP                                      & 0.141           & 2.669                             & 0.179           & 6.533                           & 0.109            & 1.310                 \\
    RTAB-Map LiDAR-I                              & \textbf{0.111}  & \textbf{1.982}                    & \textbf{0.091}  & 3.126                           & -                & -                     \\
    FAST-LIO2                                     & 0.242           & 3.633                             & 0.184           & \textbf{2.718}                  & 0.233            & \textbf{1.302}        \\
    \hline
    \hline
    \end{tabular*}

\vspace*{2pt}
    \includegraphics[width=0.91\linewidth]{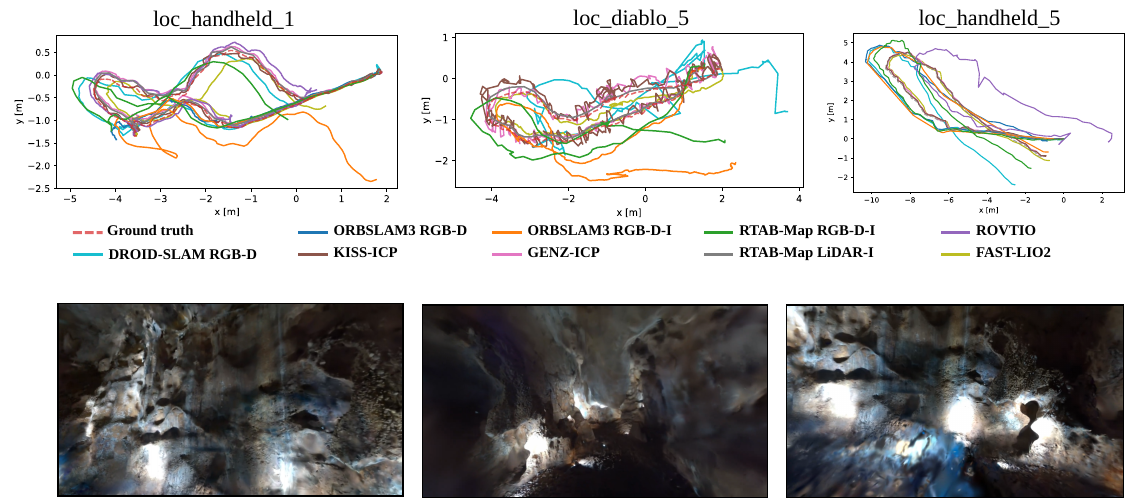}
    \captionof{figure}{In the top row, the trajectories estimated from the benchmarked algorithms are reported. For quantitative results, refer to Table~\ref{tab::sec3::results}. Three snapshots of the dense-reconstructed cave environment are shown in the bottom row. Artifacts in the reconstruction of the cave wall and floor are visible in the second and third images, respectively.}
    \label{fig::sec3::evaluation}
\vspace*{-15pt}
\end{figure*}

\subsection{SLAM and odometry}
\label{sec::3_1_evaluation_slam}

The dataset's main goal is to provide the robotics community with valuable
data to test and benchmark the trajectory accuracy of SLAM algorithms in natural
underground environments. To validate its usability, we evaluated the
performance of state-of-the-art SLAM and odometry algorithms on our dataset. For
visual SLAM, we tested ORBSLAM3 \cite{orbslam3} (RGB-D and RGB-D-I mode), RTAB-Map \cite{rtab-map} (RGB-D-I mode, with GFTT-BRISK feature
detector-descriptor), and DROID-SLAM \cite{droid-slam} (RGB-D offline mode, model pre-trained), while ROVTIO \cite{rovtio} was used to test visual-thermal-inertial odometry. ROVTIO is not a full SLAM system,
lacking any mapping and loop closure modules. For LiDAR odometry, we evaluated the performance of KISS-ICP
\cite{kiss-icp}, GENZ-ICP \cite{genz-icp}, RTAB-Map (LiDAR-I mode), and FAST-LIO2 \cite{fast-lio2}. All LiDAR pipelines perform scan-to-map ICP registration,
where point correspondences are selected using a point-to-point metric
(KISS-ICP), a point-to-plane metric (RTAB-Map, FAST-LIO2), or an adaptable metric that
combines the two (GENZ-ICP). Scans are deskewed and filtered before
registration. Time correspondences between the estimated trajectories and the
ground truth have been extracted. The paired positions have then been aligned by
estimating a rigid-body transformation between the first matched states in a
least-squares fashion \cite{metrics_scaramuzza}. We provide
trajectory accuracy metrics as Absolute Trajectory Error (ATE), computed as the
Root Mean Square Error (RMSE) of both 3D translation (ATE\textsubscript{t}) and
rotation (ATE\textsubscript{rot}). Rotation errors have been computed by
converting the pose rotation matrix to angle-axis representation and using the
rotation angle difference as the error. Table~\ref{tab::sec3::results}
summarizes the results for three selected trajectories that cover both cave rooms
and sensor modalities, while in Figure~\ref{fig::sec3::evaluation} the top views of estimated paths are displayed.

In trajectory \textit{loc\_handheld\_1}, the sensor rig is operated with a slow,
omnidirectional motion to simulate UAV movements. The cameras are mostly
pointing down, and visual features are only available where the LED headlight is
pointing. Some frames ($\sim$ 1 s) in the middle of the trajectory suffered from abrupt illumination changes due to slight movements of the headlight. In visual-only mode, tracking is
lost for ORBSLAM3 and never recovered, while tracking is maintained in visual-inertial mode, but the algorithm accumulates excessive drift that is not recovered during
the last part of the trajectory. Thermal-visual-inertial and LiDAR-based pipelines are also capable of successfully estimating the complete trajectory. In terms of passive vision, both ROVTIO and RTAB-Map outperform ORBSLAM3 for this trajectory,
and exhibit no significant issues. In trajectory \textit{loc\_diablo\_5}, the sensors are subject to more
aggressive motion, with high vibrations and fast turns. ORBSLAM3 RGB-D fails to
exit the initialization phase due to persistent high motion blur. In
visual-inertial mode, the trajectory is completed, even though it shows higher
drift that in handheld ones. Similar results are obtained with RTAB-Map RGB-D-I,
with slightly higher vertical drift than ORBSLAM3. ROVTIO estimation errors are
comparable to those of other methods for the rotational component, which is
primarily influenced by IMU data; however, translational errors are
significantly higher due to the algorithm's difficulty in tracking good features
for multiple frames. Trajectory \textit{loc\_handheld\_5} has been recorded in Sala de las Conchas,
where, as stated before, it was not possible to install the motion capture
system. To still test the algorithms, we use the estimated trajectory from
RTAB-Map in LiDAR-I mode as reference. In this sequence, larger
portions of the room are better illuminated by the light source on board. This,
together with the sensor rig's slow motion, allows a high number of visual
features to be detected and tracked. For this, all visual algorithms tested
present good trajectory accuracy, with the visual-inertial ones outperforming
RGB-D ORBSLAM3. Among visual algorithms, ROVTIO presents the highest drift, likely due to the high variance in feature depths throughout the sequence. As a monocular method, all features are initialized with a uniform depth guess; when the environment contains features at disparate distances, this fixed initialization often causes the depth estimation to converge toward local minima. We observed that DROID-SLAM RGB-D achieves competitive accuracy on positions in all the trajectories, while greater rotational errors are present. This is particularly evident in the challenging \textit{loc\_diablo\_5} sequence. The estimated trajectory drift increases towards the end of the \textit{loc\_handheld\_5} sequence, after a section where the sensor temporarily points only toward the cave floor after previously observing the wider environment. This challenging scene change appears to decrease the model's ability to maintain consistent tracking. It should be noted that DROID-SLAM was evaluated by loading the pre-trained network weights, without fine-tuning it on data from real-world subterranean environments. The monocular-only version of the model has also been tested but failed to reconstruct the trajectories in all the sequences. 

LiDAR odometry algorithms consistently outperform visual-based methods in all
scenarios. This can be mainly attributed to their insensitivity to challenging lighting conditions and to the abundance of geometrical features inside the cave rooms. In general, we observe that GENZ-ICP achieves higher accuracy than
KISS-ICP, likely due to its environment-adaptive ICP metric. For trajectory
\textit{loc\_diablo\_5}, both GENZ-ICP and KISS-ICP estimate trajectories that
closely follow the ground truth, but exhibit high oscillations around the true
path due to the rough sensor motion when installed on the DIABLO rover. Deskew is
ineffective and even decreases the estimation quality. This limitation arises
because both methods assume constant LiDAR velocity during each scan, an
assumption violated by the substantial rover's accelerations.
Conversely, RTAB-Map and FAST-LIO2 yield more stable trajectories due to the support of high-frequency IMU data: in RTAB-Map, these are used to maintain a consistent reference frame during scan deskewing, while they are tightly integrated in FAST-LIO2, where IMU measurements are pre-integrated in the state transition model of the internal Kalman filter and employed as time points during motion compensation of the scan. We note that deactivating scan feature extraction, as proposed by the authors, caused FAST-LIO2 to diverge in all the test sequences. The difficulty in tuning feature extraction parameters could explain FAST-LIO2's slightly worse position accuracy results. Generally speaking, tightly coupled sensor fusion algorithms are sensitive to time synchronization of the input data. The missing hardware synchronization in our dataset, particularly between LiDAR/RGB-D/thermal inputs and IMU measurements, could reduce the performance of the benchmarked algorithms.

\subsection{3D reconstruction and mapping}
\label{sec::3_2_mapping}

A consistent number of sequences included in the dataset allows users to test techniques for cave mapping and dense 3D reconstruction. As an example, we employed the pipeline introduced by \cite{reconstruction_alfonso} to obtain a dense reconstruction of Sala del Dosel. The method is based on the Nerfstudio Gaussian Splatting implementation Splatfacto-W \cite{Splatfacto}, and uses RGB images to
perform 3D reconstruction of the environment, with camera poses initially estimated with structure-from-motion from COLMAP \cite{colmap}. This pipeline has been chosen since it was already validated in a planetary analogue environment, and for its ease of use, since it is ROS 2 compatible. The bottom row of Figure~\ref{fig::sec3::evaluation} shows three snapshots extracted from the 3D reconstruction of the cave room. The Gaussian Splatting method successfully reconstructed the areas with sufficient illumination and multiple sensor passes, as demonstrated by the cave's left wall in the first image of Figure~\ref{fig::sec3::evaluation}. In contrast, reconstruction artifacts are visible on the right wall in the second image, where the available illumination was insufficient, and on the floor in the third image, where the sensor trajectories provided limited coverage. A video showing the complete reconstruction is provided as supplementary material.

\section{Challenges and Limitations}
\label{sec4_challenges}
Data collection in Cueva de la Victoria posed significant logistical and technical challenges. All equipment had to be introduced into the cave, transported through narrow passages, and deployed in a confined space, where a motion capture system was installed and calibrated under challenging lighting and spatial constraints. While this
enabled us to record accurate ground truth measurements, it also led to unavoidable artifacts. Motion capture hardware and cabling are visible in the background of part of these sequences, and heat dissipated by the cameras is visible in the thermal images, as shown in Figure \ref{fig::sec2::example_data}. In addition, operators occasionally appear in the rear portion of some LiDAR scans due to the need for handheld operation and tethered data logging. To preserve the integrity of the natural environment, our dataset includes sequences captured without instrumentation in the scene, albeit without ground truth.

\section{Conclusions}
\label{sec5_conclusions}
This article presents a multimodal dataset collected in two distinct rooms of a natural
karstic cave, designed for evaluating SLAM methods for underground autonomous
systems. The dataset, which has been released to the public, includes visual data (RGB, depth, IR), LiDAR scans, and IMU measurements, along with ground truth 6D sensor poses and velocities obtained from a
motion capture system installed directly inside the cave. Benchmarking with state-of-the-art
SLAM algorithms demonstrates the quality and practical usability of the dataset. Future extensions will focus on including complementary sensing modalities, such as stereo and event-based cameras, and higher-precision IMUs. We also aim to acquire data under more challenging conditions, including the presence of dust in the air, to better reflect operational scenarios encountered on exploration missions. Finally, we plan to carry out a longer data collection campaign that will allow us to instrument multiple cave rooms with the motion capture system.





\section*{ACKNOWLEDGMENT}

This work is part of a Piano Nazionale di Ripresa e Resilienza PhD
scholarship funded by the Italian Ministry of Education, University and
Research – DM 352/2022 and part of the project INSIGHT (PID2024-160373OB-C21) funded by MICIU / AEI / 10.13039/501100011033 / FEDER, UE. It was made possible thanks to the project ``Proyecto de actividad arqueológica en las cuevas de la Victoria e Higuerón Tesoro (Rincón de la Victoria, Málaga, España). Estudio de arte rupestre y sondeos geoarqueológicos'' authorized by the Regional Ministry of Culture of the Andalusian Government and the Rincón de la Victoria City Council. We extend our sincere appreciation to the project's director, María del Mar Espejo Herrerías, and especially to Pedro Cantalejo Duarte and Pedro Cantalejo Espejo for granting us access to Cueva de la Victoria. We would also like to thank Giovanni Mastrorocco, Levin Gerdes, and Jesús Juli Fernández for their invaluable help.

\bibliographystyle{IEEEtran}
\bibliography{IEEEabrv,bibliography.bib}

\end{document}